\documentclass[sigconf]{acmart}

\AtBeginDocument{%
  \providecommand\BibTeX{{%
    \normalfont B\kern-0.5em{\scshape i\kern-0.25em b}\kern-0.8em\TeX}}}

\setcopyright{rightsretained}
\acmConference[Preprint]{}{arXiv}{2021}
\acmDOI{}
\acmISBN{}

\usepackage{soul,longtable,lscape} 
\usepackage{xcolor,hyperref}
\usepackage{subfig}
\usepackage{dsfont}

\begin{document}

\title{Emergence of Structural Bias in Differential Evolution}

\author{Bas van Stein}
\authornote{Both authors contributed equally to this research.}
\orcid{0000-0002-0013-7969}
\affiliation{%
  \institution{LIACS, Leiden University}
  \country{The Netherlands}}
  \email{b.van.stein@liacs.leidenuniv.nl}
\author{Fabio Caraffini}
\authornotemark[1]
\authornote{Corresponding author}
\orcid{0001-9199-7368} 
\affiliation{
  \institution{Institute of Artificial Intelligence, \\De Montfort University}
  \country{Leicester, UK}}
\email{fabio.caraffini@dmu.ac.uk}

\author{Anna V. Kononova}
\orcid{0002-4138-7024}
\affiliation{
  \institution{LIACS, Leiden University}
  \country{The Netherlands}}
\email{a.kononova@liacs.leidenuniv.nl}

\renewcommand{\shortauthors}{van Stein and Caraffini, et al.}

\begin{abstract}
Heuristic optimisation algorithms are in high demand due to the overwhelming amount of complex optimisation problems that need to be solved. The complexity of these problems is well beyond the boundaries of applicability of exact optimisation algorithms and therefore require modern heuristics to find feasible solutions quickly. These heuristics and their effects are almost always evaluated and explained by particular problem instances. In previous works, it has been shown that many such algorithms show structural bias, by either being attracted to a certain region of the search space or by consistently avoiding regions of the search space, on a special test function designed to ensure uniform `exploration' of the domain. In this paper, we analyse the emergence of such structural bias for Differential Evolution (DE) configurations and, specifically, the effect of different mutation, crossover and correction strategies. We also analyse the emergence of the structural bias during the run-time of each algorithm. We conclude with recommendations of which configurations should be avoided in order to run DE unbiased.
\end{abstract}



\maketitle

\section{Introduction}\label{sect:intro}

Heuristic optimisation algorithms are in high demand in the modern world due to the overwhelming amount of optimisation problems that need to be solved to sustain the ongoing technological boom. Such problems grow not only in their amount but also in their complexity --- well beyond the boundaries of applicability of exact optimisation algorithms. Luckily, modern heuristics can deliver (strictly speaking, sub-optimal) solutions of sufficiently good quality, if designed and tuned appropriately. The heuristic optimisation community is yet to build the underlying theory/methodology for such efficient design and tuning process, but first steps are already taken \cite{fruhwirth1998theory,leguizamBoundarySerch,Lehre2012,HansterK2017flaccogui,KerschkeT2019flacco}. Characterisation of behaviour of heuristic algorithms, studied in this publication, falls within such methodological aspiration. 

Effect of the application of some algorithm is always observed on a particular problem or a collection of problems, and success or failure of some algorithm typically is `explained' by features of the problem \cite{mersmann2011exploratory}. Algorithms `specialise' on some problems more than on others\footnote{This is perfectly in line with the No Free Lunch Theorem (NFLT) \cite{NFLT} that roughly states that performance of all algorithms is the same if averaged over all possible objective functions. In other words, no best optimiser exists.}. Would it be reasonable to assume that this happens due to some feature of the algorithm that makes it more or less `predisposed' to some (kind of) problems? How could such features be studied since, as stated above, an algorithm cannot be examined on its own but only applied to some problem\footnote{This statement is clearly not fully true since an algorithm can be studied theoretically. However, most modern heuristics cannot be subject to such analysis without major simplifications.}?

A step in this direction has been made in \cite{Kononova2015} where a concept of the so-called structural bias (SB) has been introduced in relation to the characterisation of population-based heuristic optimisation algorithms. In such algorithms, a set of operators is applied to a collection of sampled points (population) in an iterative manner. Points `move' inside the domain driven by some selection criteria, based on the survival-of-the-fittest analogy. The authors of \cite{Kononova2015} argue that the iterative nature of the application of a limited number of operators responsible for generation and selection of new sampled points and their interplay, can lead to the emergence of artificial `biases' that interfere with the direction of sampling for the new points, regardless of the problem/objective function. Such theoretical possibility appears more than plausible if population-based algorithms are contrasted with Iterated Function Systems (IFS) \cite{Barnsley1988} with its numerous results \cite{DAniello2017,DAniello2019} like the collage theorem which states that for every possible image there exists a strictly contractive IFS whose attractor arbitrarily closely approximates this image. 

To identify structural bias, a special $f_0$ test function has been proposed \cite{Kononova2015} which allows decoupling behaviour of the algorithm from the objective function by assigning independent uniformly distributed random numbers instead of objective function values. Best solutions found by the optimisation heuristic in a series of runs on such objective function would then naturally follow some distribution: a structurally unbiased algorithm would result in a uniform distribution of final points, meanwhile structurally biased algorithms would show some `preference' to part(s) of the domain, i.e. not return the equiprobable uniform distribution. 

The concept of structural bias has been successfully investigated for a large number of algorithms and speculations have been made regarding possible mechanisms of its formation in different frameworks such as GA \cite{Kononova2015} where SB becomes more severe with increasing population size and PSO \cite{Kononova2015} where SB appears to be minimised for medium-sized populations. The concept has been further applied to population-free optimisation heuristics such as single solution methods \cite{Kononova2020CEC} and specific versions of Estimation of Distribution Algorithms (EDA) \cite{Kononova2020PPSN} --- both have been found to possess significant amounts of SB. 

Differential evolution (DE) \cite{Storn1995} is one of the most popular continuous heuristic optimisation methods. Apart from its well-known advantages such as a small number of parameters and robust performance for a wide range of problems \cite{DEeng,bib:Miettinen99}, it has been previously shown \cite{Caraffini2019} to possess no structural bias for the majority of widely used configurations (for a fixed parameter setting). In this paper, we extend such study into a complete investigation of the \textit{emergence of structural bias} in DE from two points of view: (1) for which values of parameters DE configurations become biased, (2) at what point in time do runs of a DE configuration become biased and how such bias evolves. We call this the \emph{emergence} of bias, as each algorithm configuration starts with an unbiased initial population, according to most general specifications of DE\footnote{In fact, most population-based heuristic optimisation algorithms have uniformly distributed initial populations.}. Moreover, we investigate the strength of SB for DE configurations with recommended parameter settings.

The structure of this paper is as follows. In Section \ref{section:Method} the statistical test that is used to measure structural bias is explained and all the DE variations and configurations analysed in this study are defined. Next, the experimental setup is described in detail. In Section \ref{sect:emergence_param}, the emergence of bias over different configurations in parameter space is analysed and discussed. In Section \ref{sect:emergence_time}, the analysis of structural bias for a selection of strongly biased configurations is extended to evaluate how the bias evolves over evaluations of the algorithm. Finally, we conclude our observations and mention future work directions in Section \ref{sect:conclusion}.

\section{Methodology for measuring bias}
\label{section:Method}
\subsection{Statistical test for structural bias}\label{sec:stats-measure}
Previous work~\cite{Kononova2020CEC,Kononova2020PPSN} has investigated the structural bias for a wide range of optimisation algorithms on a theoretical function $f_0$, where (local) optima are distributed uniformly in its domain:
\begin{equation}
    \text{minimise } f_0:[0,1]^n\to[0,1], \text{ where } \forall x, f_0(x) \sim \mathcal{U}(0,1).
\end{equation}
Despite its simplicity, this function is ideal for studying structural bias since it imposes no selection pressure on algorithms. Consequently the non-uniformity of the optimisation outcome, if it were to occur, must be attributable to the structural bias of the algorithm. Based on this consideration, we attempted to investigate SB by means of visual and statistical tests~\cite{Kononova2020CEC,Kononova2020PPSN}. The former entails plotting in a parallel coordinate chart the final points from multiple independent runs, which the researchers examine visually, while the latter checks for each dimension if the final points follow the uniform distribution $\mathcal{U}(0,1)$ through the well-known Anderson-Darling (AD) test. Despite its natural interpretability, the visual test is highly subjective and cannot be automated to check results from large experiments. Meanwhile, the statistical approach suffers from a potentially low statistical power when the sample size is not large enough, and it also yields considerable inconsistencies to the visual test on the cases deemed to be of no or mild bias~\cite{Kononova2020PPSN}.

Due to the substantial number of DE configurations to investigate, we decided to use the statistical approach since applying the visual test here would be extremely strenuous and impractical. For the sake of being self-contained, we quickly recap~\cite{Kononova2020PPSN} the details of the statistical measure before proceeding to the experimental setup: given the final points $\{\mathbf{x}^{(1)},\mathbf{x}^{(2)}\ldots, x^{(r)}\}\subset [0,1]^n$ found by an algorithm in $r=600$ independent runs\footnote{This sample size is deemed sufficient as the corresponding AD test yields a statistical power of 1 when simulated under a mixture of Beta distributions. Such relatively large sample size is necessary as shown in \cite{Vermetten2021_anisotropy}.} on $f_0$, we apply an AD test on each component/dimension of those final points against $\mathcal{U}(0,1)$, resulting in a collection of test statistics $\{A^2_1, A^2_2,\ldots, A^2_n\}$ and p-values $\{p_1, p_2, \ldots, p_n\}$, where $A^2_i = \int_0^1 (\widehat{F}_{i}(t) - t)^2/t(1-t) \mathrm{d} t$ for $i=1, 2,\ldots, n$ and $\widehat{F}_{i}(t)$ is the empirical cumulative distribution function on the $i$th component. After adjusting the p-values with the so-called Benjamini–Yekutieli method~\cite{BenjaminiY01} (which is a common means for handling the multiple comparison problem), we proceed to aggregate all $A^2$ statistics; this leads to significance decisions for quantifying the overall structural bias, namely $\operatorname{SB} = \frac{1}{n}\sum_{i=1}^n A_i^2 \mathds{1}(p_i^{\text{adjust}} \leq \alpha)$, where $p_i^{\text{adjust}}, \mathds{1}$, and $\alpha$ are the adjusted p-values, the characteristic function, and user-specified significance level, respectively.

\subsection{Considered algorithms}\label{sect:algorithms}

In DE jargon, the notation DE/\texttt{x}/\texttt{y}/\texttt{z} is used to fully describe the DE variant under consideration. Due to the simplicity of the DE algorithmic framework, see \cite{Storn1995,DasS11,Caraffini2019,Das2016} for details, it is indeed sufficient to indicate a mutation operator \texttt{x}, the number of the so-called `difference vectors' it is expected to employ, and a crossover operator to allow for its implementation. Usually overlooked, or superficially assumed to be an irrelevant algorithmic detail, the employed Strategy of Dealing with Infeasible Solutions (SDISs) is also a key aspect that should be indicated to complete the algorithm description within an experimental context --- as consistently suggested by recent studies on DE \cite{Caraffini2019lego,Caraffini2019,Kononova2020_outside} and heuristic optimisation algorithms in general \cite{Kononova2020CEC,Kononova2020PPSN}. 

\subsubsection{Mutation}
The role of the mutation operator in DE is to linearly combine distinct individuals from the population to generate a mutant vector. The most established mutation operators \texttt{x} have self-explanatory names, referring to the direction where the obtained mutant vector is expected to point (e.g. towards the current best individual, towards a random one, from the a current individual to a random one, etc.). Note that this direction is altered by adding, e.g. to the best individual, or to the vector from the current individual to the best individual, etc., at least one difference vector --- their number is denoted with \texttt{y} in each DE variant. Each such difference vector is obtained by taking the scaled difference (parameter $F$ being the scaling coefficient) of two distinct randomly chosen individuals from the population. 

For this study, we have selected $7$ widely used \texttt{x}/\texttt{y} combinations:  \texttt{rand/1}; \texttt{rand/2}; \texttt{best/1}; \texttt{best/2}; \texttt{rand-to-best/2}; \texttt{current-to- best/1}; \texttt{current-to-rand/1}; whose implementation details and code are available in e.g. \cite{DasS11,Das2016} and \cite{CaraffiniSOS_2020, mendeley2021emergence} respectively.

\subsubsection{Crossover}
To complete a generation cycle in any DE variant, a mutant must be produced for each individual in the population so that crossover can be applied, i.e. between the current individual and its mutant, to generate corresponding offspring solutions. Hence, a generation cycle requires a number of fitness function evaluations equal to the population size \texttt{p}. Amongst the proposed crossover logics, the binary (i.e. \texttt{z = bin}) and exponential (i.e. \texttt{z = exp}) operators are the most established in the DE world. Note that both \texttt{exp} and \texttt{bin} require only one parameter to function, i.e. the crossover rate $C_r$, but behave very differently when employed with similar $C_r$ values. Considerations on the behaviour of the \texttt{exp} and \texttt{bin} operators and corresponding pseudocodes are available in \cite{Caraffini2019,Kononova2020_outside}. For these reasons, both the operators are included in our experimental setup. 

It must be pointed out that the \texttt{current-to-rand/1} mutation does not require an additional crossover strategy as it internally performs recombination between involved individuals \cite{Caraffini2019,Das2016}. Therefore, by combining the employed mutation and crossover operators we form the following $13$ DE variants for our investigation:
\begin{itemize}
\item \texttt{DE/rand/1/bin} and \texttt{DE/rand/1/exp};
    \item \texttt{DE/rand/2/bin} and \texttt{DE/rand/2/exp};
    \item \texttt{DE/best/1/bin} and \texttt{DE/best/1/exp};
    \item \texttt{DE/best/2/bin} and \texttt{DE/best/2/exp};
    \item \texttt{DE/current-to-best/1/bin} and \\ \texttt{DE/current-to-best/1/exp};
    \item \texttt{DE/rand-to-best/2/bin} and \texttt{DE/rand-to-best/2/exp}; 
    \item \texttt{DE/current-to-rand/1}.
 \end{itemize}

For the sake of clarity, it is worth mentioning that each offspring solution competes with the individual generating them after each generation, thus forming a new population for the following iteration cycle.

\subsubsection{SDIS}
The choice of the SDISs is of high importance, in particular for highly multidimensional problems, as it is more likely to generate infeasible solutions \cite{Kononova2020_outside}. Therefore, we equip each \texttt{DE/x/y/z} algorithm under investigation with the $6$ SDISs below:  
\begin{itemize}
    \item Complete one-sided truncated normal strategy denoted as \texttt{COTN};
    \item dismiss strategy denoted here as \texttt{dis};
    \item mirror strategy denoted here as \texttt{mir}; 
    \item saturation strategy denoted here as \texttt{sat};
    \item toroidal strategy denoted here as \texttt{tor};
    \item uniform strategy denoted here as \texttt{uni}.
\end{itemize}
Note that the selected SDISs feature different working logics, as suggested by their self-explanatory names. For a detailed description we recommend \cite{mendeley2021emergence,Kononova2020PPSN,Kononova2020_outside}.

\subsection{Experimental setup}\label{sect:setup}

To perform a thorough analysis, we have discretised the DE parameter space as follows:
\begin{itemize}
    \item population size $p\in\{$5$, $20$, $100$\}$;
    \item scale factor $F\in \\ \{0.05,0.266,0.483,0.7,0.916,1.133,1.350, 1.566,1.783, 2.0\}$;
    \item crossover rate $C_r \in\{0.05,0.285,0.52,0.755,0.99\}$.
\end{itemize}

In this paper, we are studying the total of $3\cdot6\cdot10$+$12\cdot3\cdot6\cdot10\cdot5=10980$ algorithm configurations since:
\begin{itemize}
    \item \texttt{DE/current-to-rand/1} is used with $3$  population sizes, $6$ SDISs and $10$ values for the scale factor F;
    \item the remaining $12$ DE variants are used with $3$ population sizes, $6$ SDISs, $10$ values for the scale factor $F$ and $5$ values for the crossover rate $C_r$.
\end{itemize}

To have a statistically significant sample size, we have executed each one of these $10980$ configurations $600$ times with a computational budget of $10000\cdot n=300000$ fitness function evaluations per run. This experimentation has been performed in the SOS platform \cite{CaraffiniSOS_2020} --- a link to the source code of the entire experimental setup is provided in \cite{mendeley2021emergence}. 

Note that when results of the DE variants under study --- used with a specific population size and equipped with a specific SDIS --- are graphically reported in the $F$-$C_r$ space later on in this paper, their names follow notation \texttt{DE/x/y/z-pN-SDIS}, where \texttt{N} indicates the value assumed by \texttt{p}.

\begin{figure}
\centering
\includegraphics[width=.35\textwidth,trim=0mm 3mm 1mm 1mm,clip]{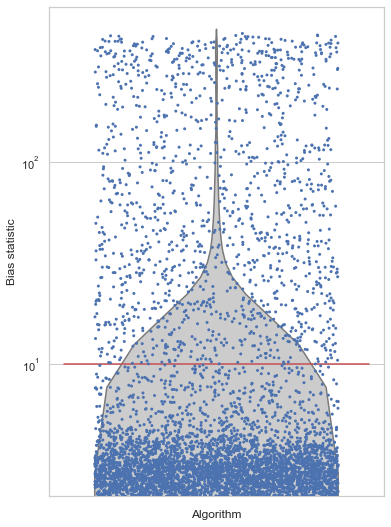}
\caption{Distribution of bias statistic over all methods and parameter settings considered in this study, with the recommended thresholds: $[0,0]$ no SB, $(0,10]$ mild SB and $(10,\infty)$ strong SB. For examples of positions of final points for these three cases, see Figures~\ref{fig:pc_noSB},~\ref{fig:pc_mildSB} and ~\ref{fig:pc_strongSB}, respectively.}\label{fig:distributionBias}
\end{figure}

\begin{figure*}
\centering
\subfloat[][\texttt{DE/best/1/bin-p20-COTN} with\newline $F=0.92$, $C_r=0.05$: no structural bias ]{\includegraphics[width=.33\textwidth,trim=25mm 5mm 25mm 10mm,clip]{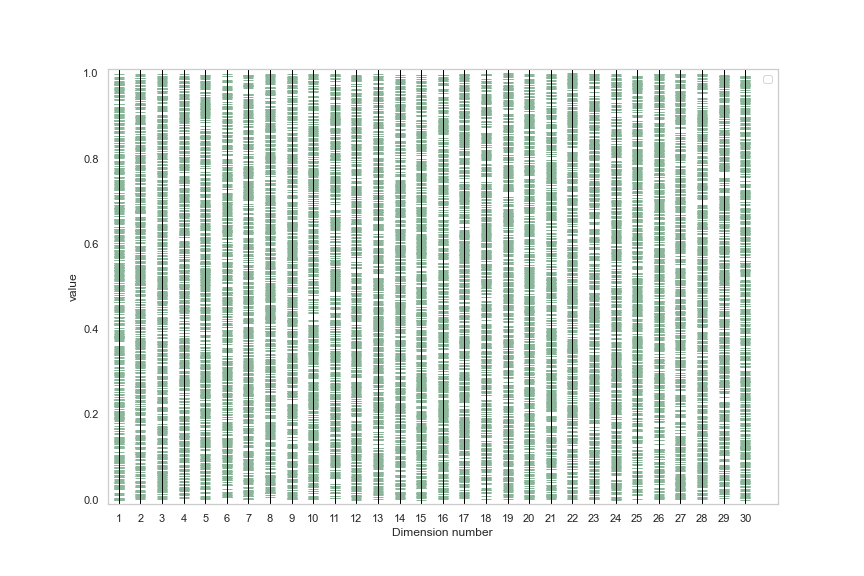}\label{fig:pc_noSB}}
\subfloat[][\texttt{DE/curr-to-best/1/bin-p20-uni} with \newline $F=0.7$, $C_r=0.76$: mild structural bias to centre]{\includegraphics[width=.33\textwidth,trim=25mm 5mm 20mm 10mm,clip]{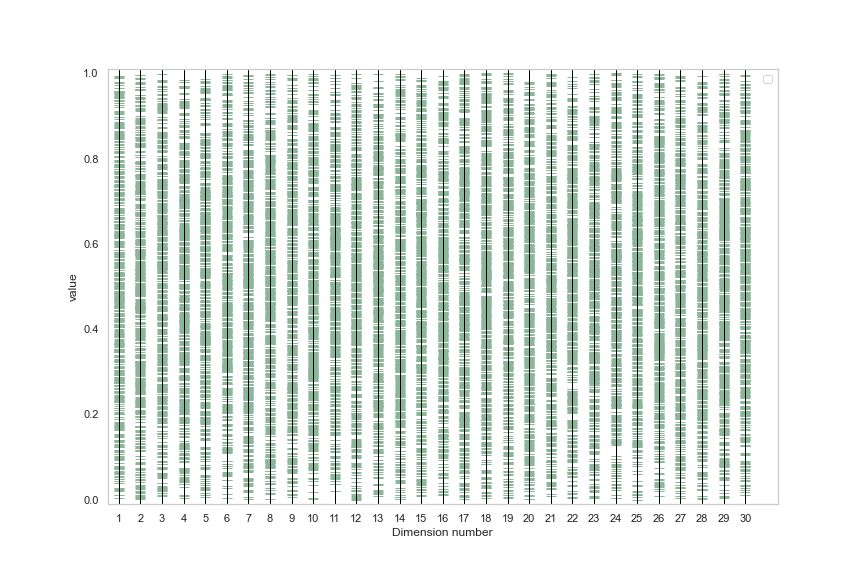}\label{fig:pc_mildSB}}
\subfloat[][\texttt{DE/curr-to-rand/1-p100-sat} with \newline $F=0.05$: strong structural bias to centre]{\includegraphics[width=.33\textwidth,trim=25mm 5mm 20mm 10mm,clip]{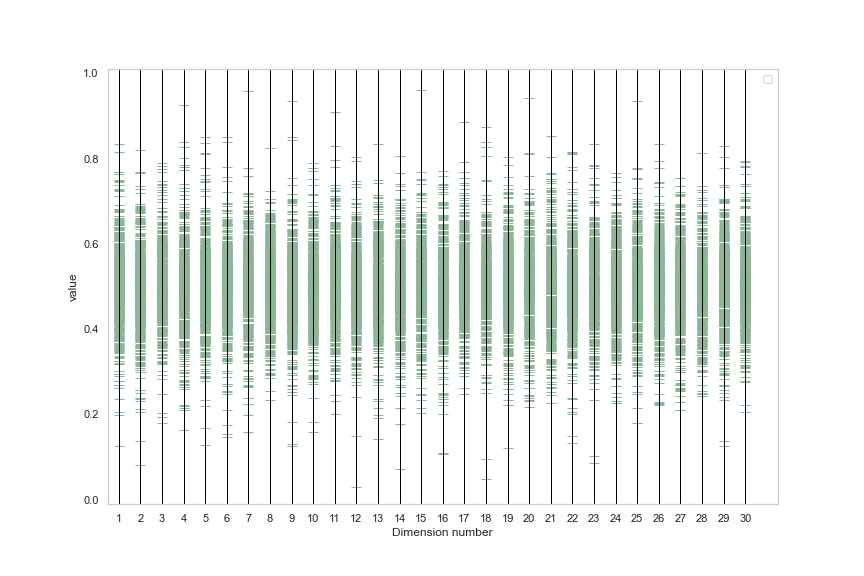}\label{fig:pc_strongSB}}
\caption{Examples of visible bias shown in parallel coordinate plots of the final results from $600$ runs of configurations with given parameter setting.}\label{fig:examplesBias}
\end{figure*}

\section{Emergence of structural bias in parameter space}\label{sect:emergence_param}

Out of 10980 considered algorithm configurations, 4737 have showed at least some bias ($0 < SB < 10$) and 1343 algorithm configurations have showed strong bias ($SB \geq 10$), see Figure~\ref{fig:distributionBias}. Moreover, the kind of SB observed varies: the most obvious cases showing bias to the centre of the search domain and bias to the edges of the search domain (Figure \ref{fig:examplesBias}).

The influence of the DE parameters, $F, C_r$ and $N$, is analysed by looking at the trends in the bias indicator for different crossover, mutation and correction strategies as shown in Figure \ref{fig:combiFCr}.

\begin{figure*}
\centering
\subfloat[][\texttt{DE/best/1/bin-p5-sat}]{\includegraphics[width=.25\textwidth,trim=55mm 10mm 20mm 33mm,clip]{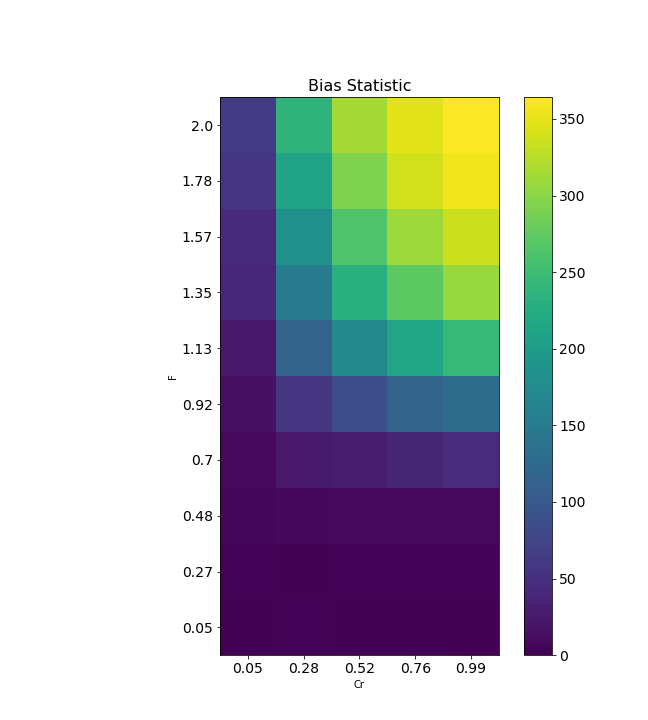}\label{fig:DEbobsP5}}
\subfloat[][\texttt{DE/best/1/bin-p20-sat}]{\includegraphics[width=.25\textwidth,trim=55mm 10mm 20mm 33mm,clip]{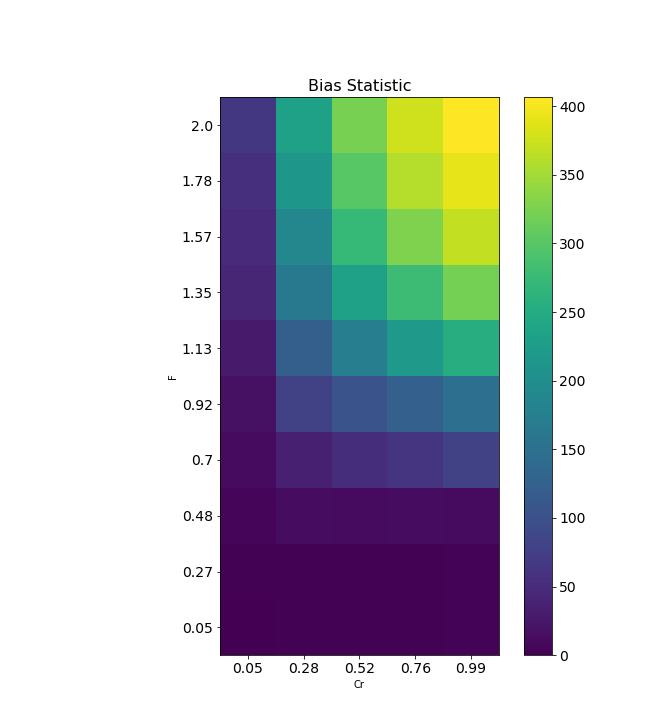}\label{fig:DEbobsP20}}
\subfloat[][\texttt{DE/best/1/bin-p100-sat}]{\includegraphics[width=.25\textwidth,trim=55mm 10mm 20mm 33mm,clip]{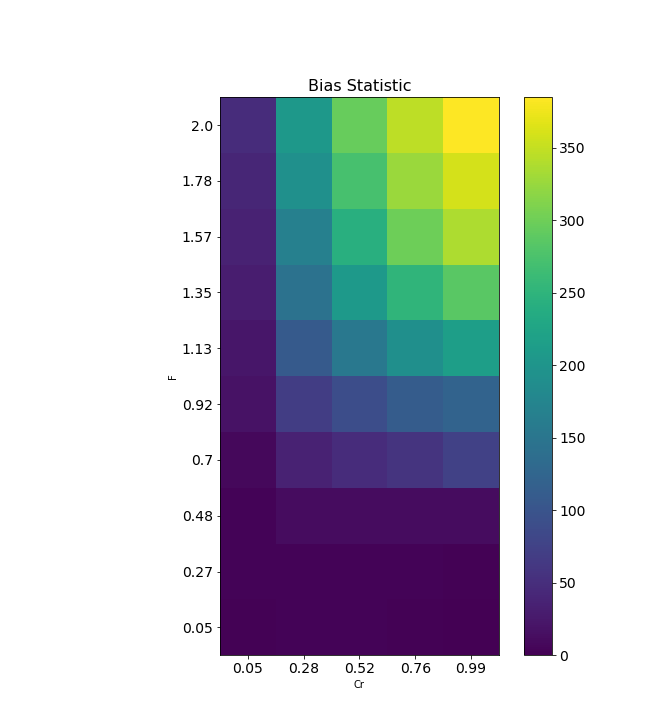}\label{fig:DEbobsP100}} 
\subfloat[][\texttt{DE/best/1/exp-p100-sat}]{\includegraphics[width=.25\textwidth,trim=55mm 10mm 20mm 33mm,clip]{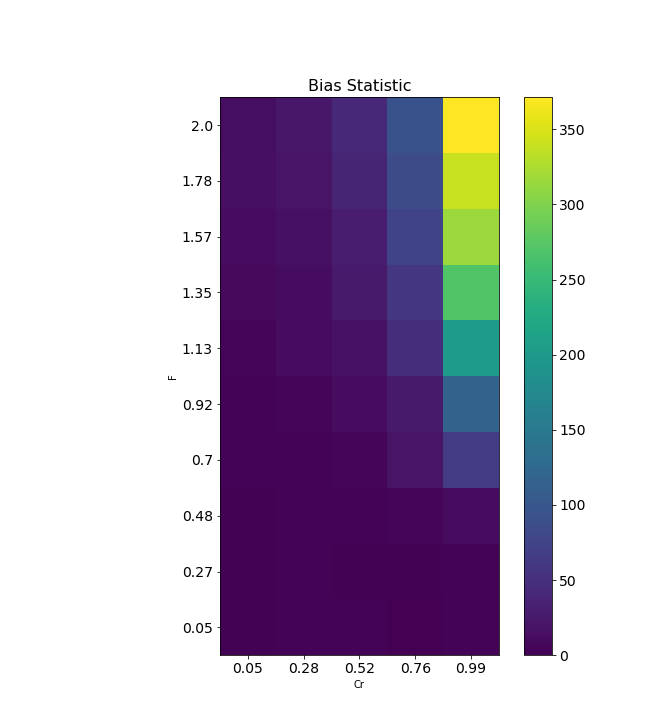}\label{fig:DEboesP100}}\\ 
\vspace{-3mm}
\subfloat[][\texttt{DE/best/1/bin-p5-tor}]{\includegraphics[width=.25\textwidth,trim=55mm 10mm 20mm 33mm,clip]{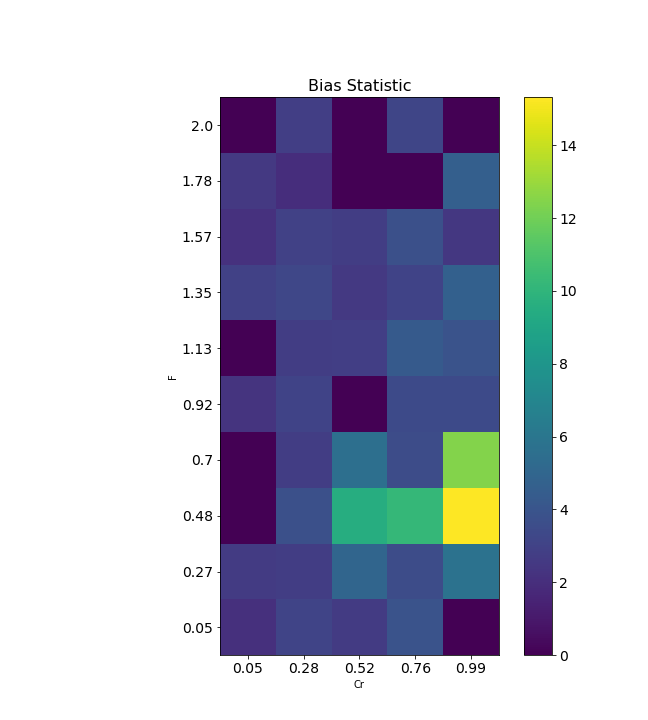}\label{fig:DEbobtP5}}
\subfloat[][\texttt{DE/best/1/bin-p20-uni}]{\includegraphics[width=.25\textwidth,trim=55mm 10mm 20mm 33mm,clip]{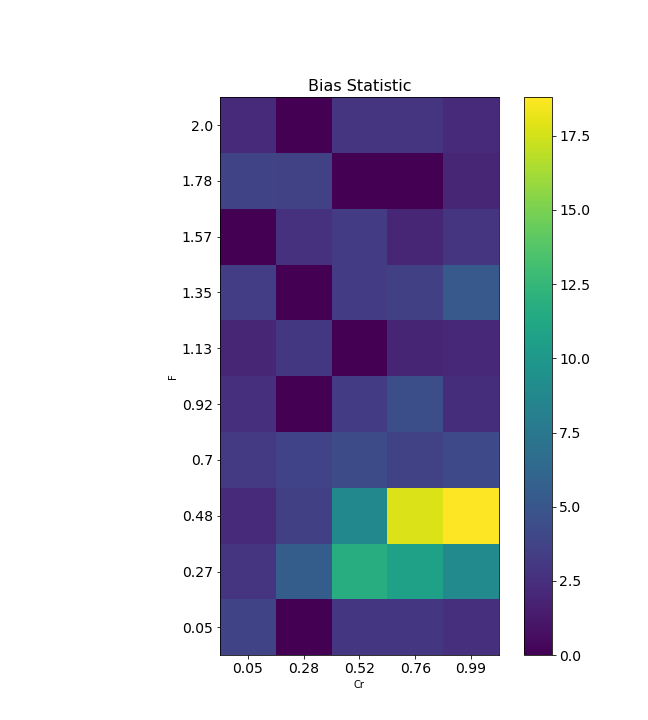}\label{fig:DEbobuP20}}
\subfloat[][\texttt{DE/best/1/bin-p100-dis}]{\includegraphics[width=.25\textwidth,trim=55mm 10mm 20mm 33mm,clip]{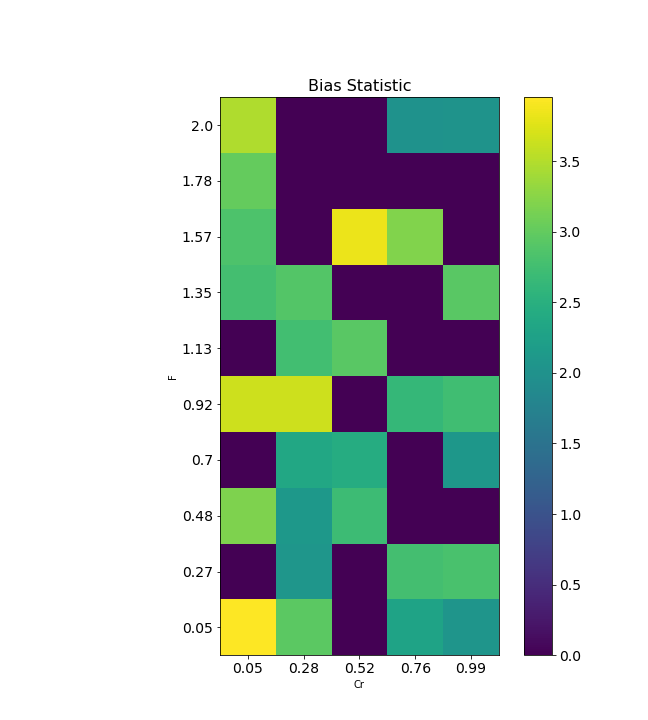}\label{fig:DEbobd_P100}}
\subfloat[][\texttt{DE/curr-to-best/1/exp-p5-sat}]{\includegraphics[width=.25\textwidth,trim=55mm 10mm 20mm 33mm,clip]{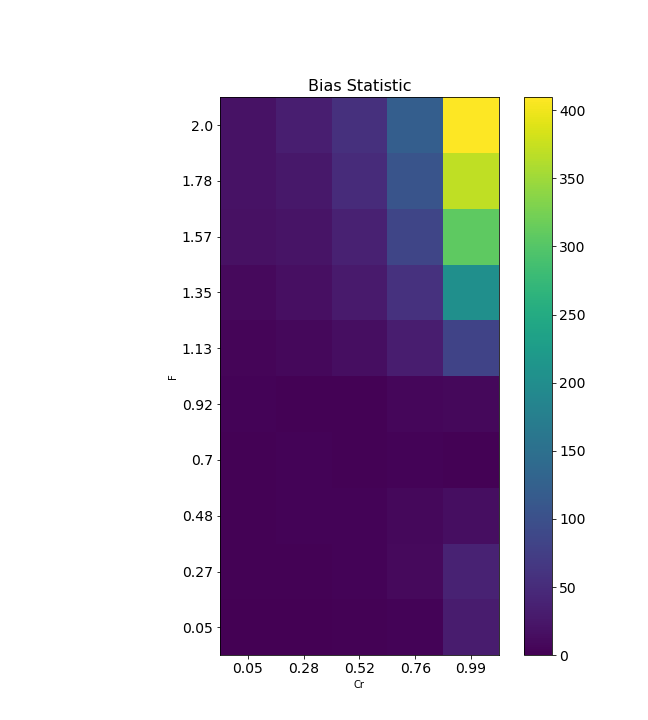}\label{fig:DEctboes_P5}}\\ 
\vspace{-3mm}
\subfloat[][\texttt{DE/best/2/bin-p20-COTN}]{\includegraphics[width=.25\textwidth,trim=55mm 10mm 20mm 33mm,clip]{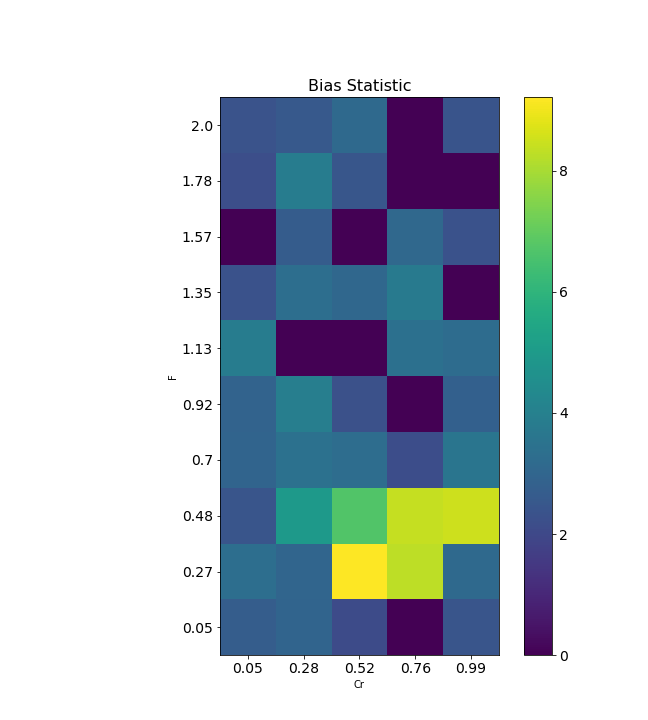}\label{fig:DEbtbc_P20}}
\subfloat[][\texttt{DE/rand-to-best/2/bin-p5-COTN}]{\includegraphics[width=.25\textwidth,trim=55mm 10mm 20mm 33mm,clip]{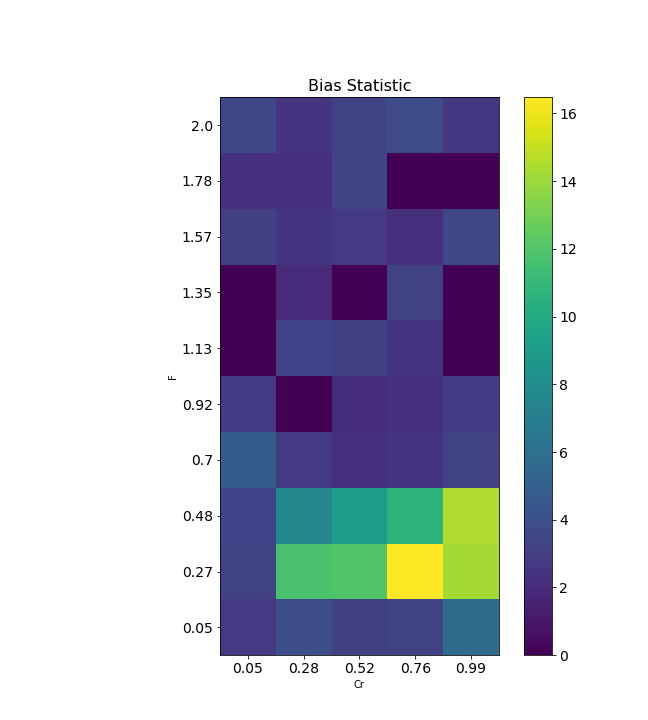}\label{fig:DErtbtbc_P5}}
\subfloat[][\texttt{DE/curr-to-best/1/exp-p100-uni}]{\includegraphics[width=.25\textwidth,trim=55mm 10mm 20mm 33mm,clip]{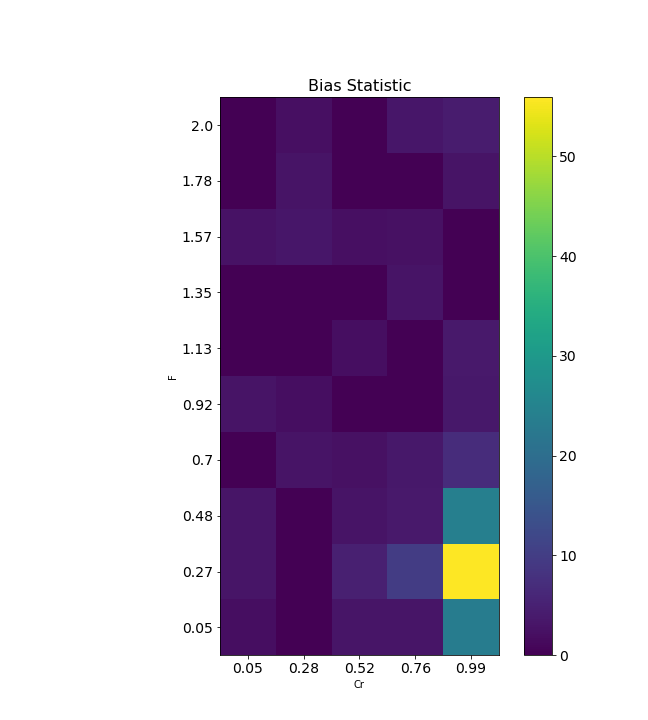}\label{fig:DEctboeu_P100}}
\subfloat[][\texttt{DE/rand-to-best/2/bin-p20-tor}]{\includegraphics[width=.25\textwidth,trim=55mm 10mm 20mm 33mm,clip]{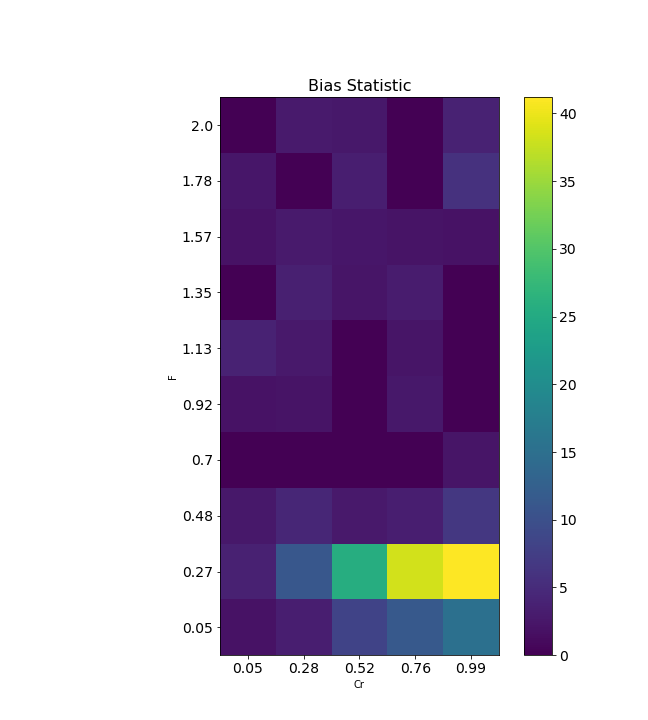}\label{fig:DErtbtbt_P20}}
\caption{Bias indicators for different algorithms and population sizes for $F$ and $C_r$ settings. Brighter values indicate higher bias. Colour scale is not uniform across figures.}
\end{figure*}

\begin{figure*}
\centering
\subfloat[][\texttt{p20-COTN}]{\includegraphics[width=.10\textwidth,trim=140mm 22mm 20mm 33mm,clip]{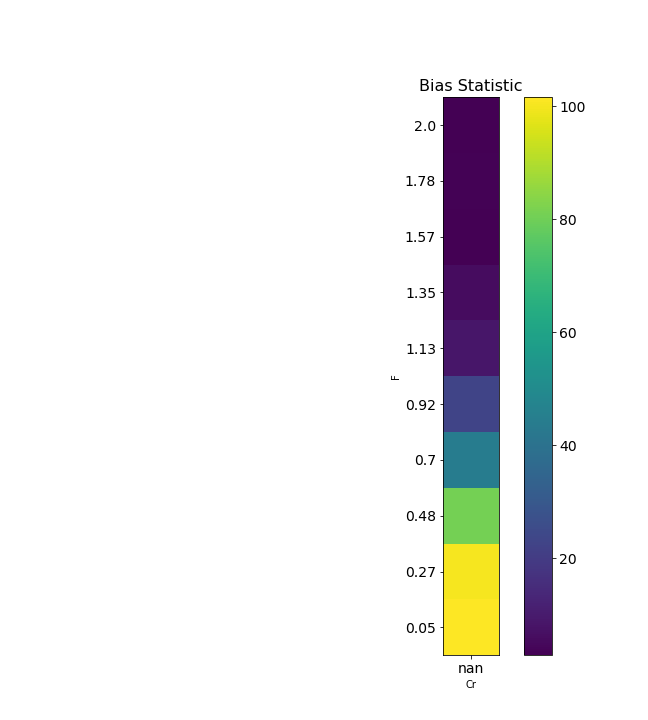}}
\subfloat[][\texttt{p20-dis}]{\includegraphics[width=.10\textwidth,trim=140mm 22mm 20mm 33mm,clip]{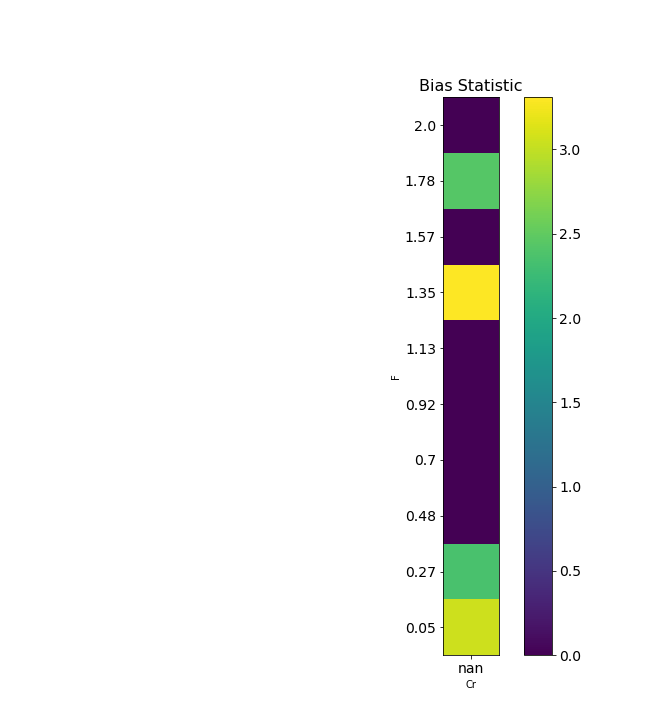}}
\subfloat[][\texttt{p20-mir}]{\includegraphics[width=.10\textwidth,trim=140mm 22mm 20mm 33mm,clip]{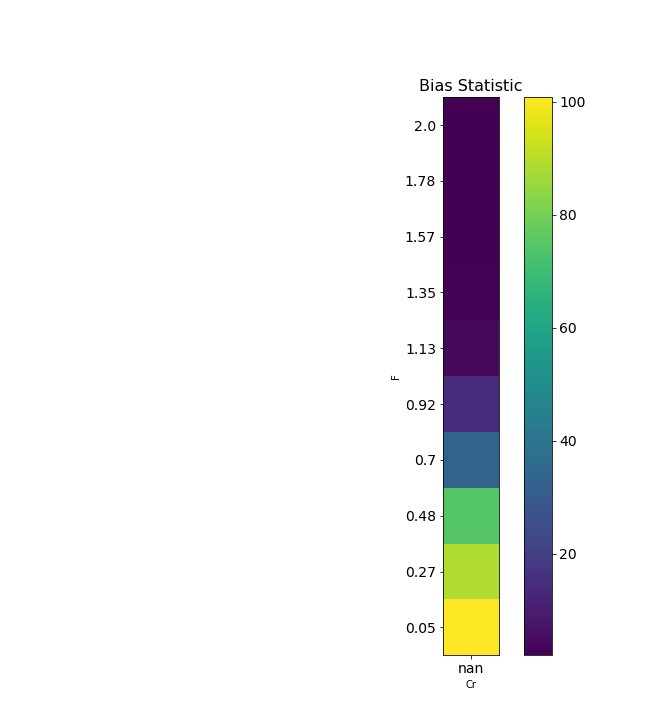}}
\subfloat[][\texttt{p20-sat}]{\includegraphics[width=.10\textwidth,trim=140mm 22mm 20mm 33mm,clip]{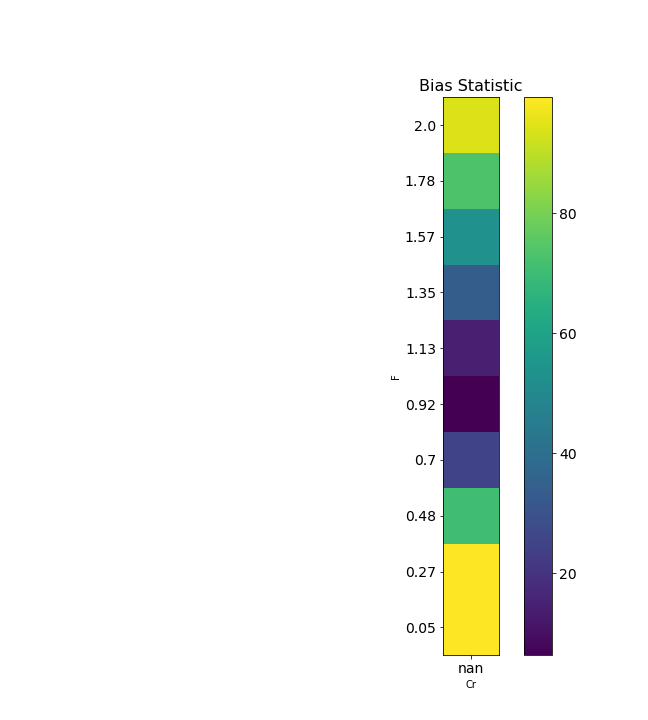}\label{fig:p20sat}}
\subfloat[][\texttt{p20-tor}]{\includegraphics[width=.10\textwidth,trim=140mm 22mm 20mm 33mm,clip]{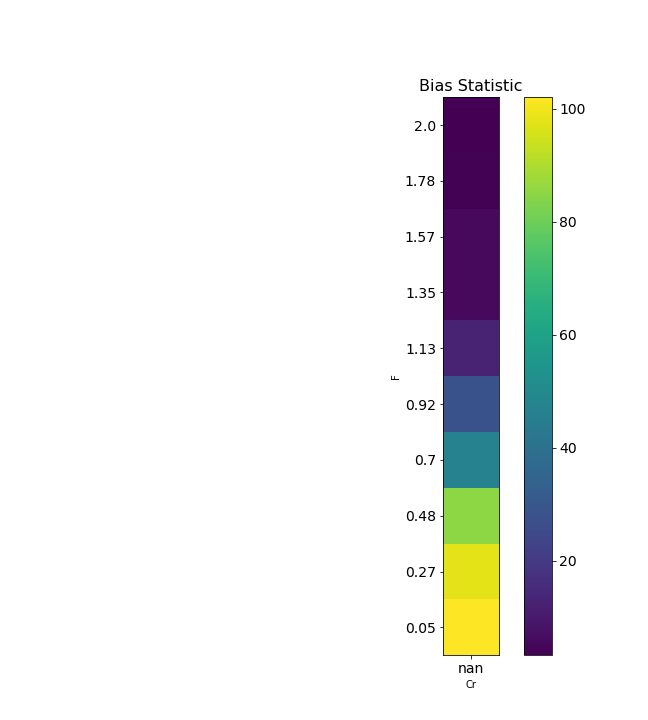}}
\subfloat[][\texttt{p100-COTN}]{\includegraphics[width=.10\textwidth,trim=140mm 22mm 20mm 33mm,clip]{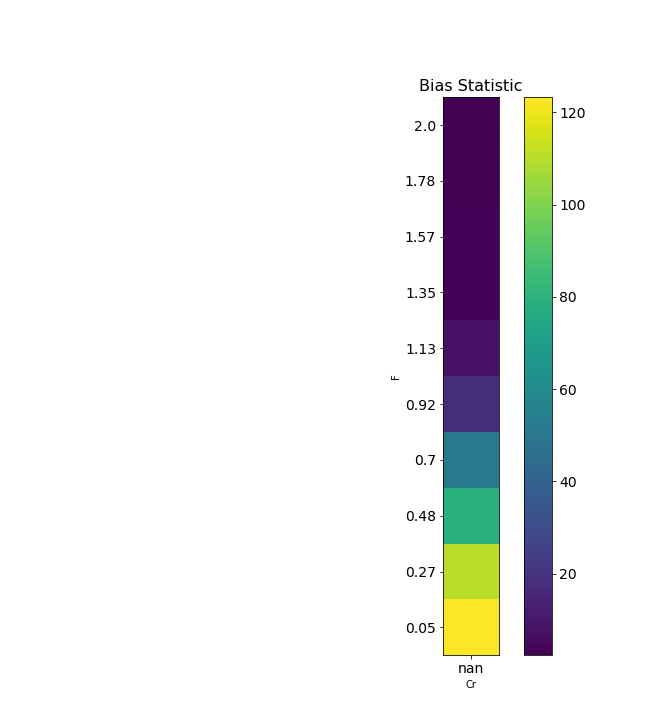}}
\subfloat[][\texttt{p100-dis}]{\includegraphics[width=.10\textwidth,trim=140mm 22mm 20mm 33mm,clip]{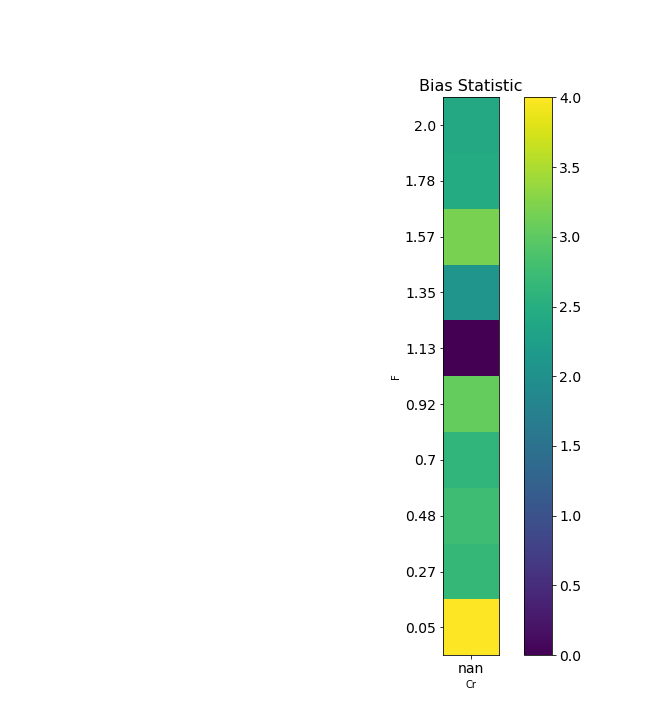}}
\subfloat[][\texttt{p100-mir}]{\includegraphics[width=.10\textwidth,trim=140mm 22mm 20mm 33mm,clip]{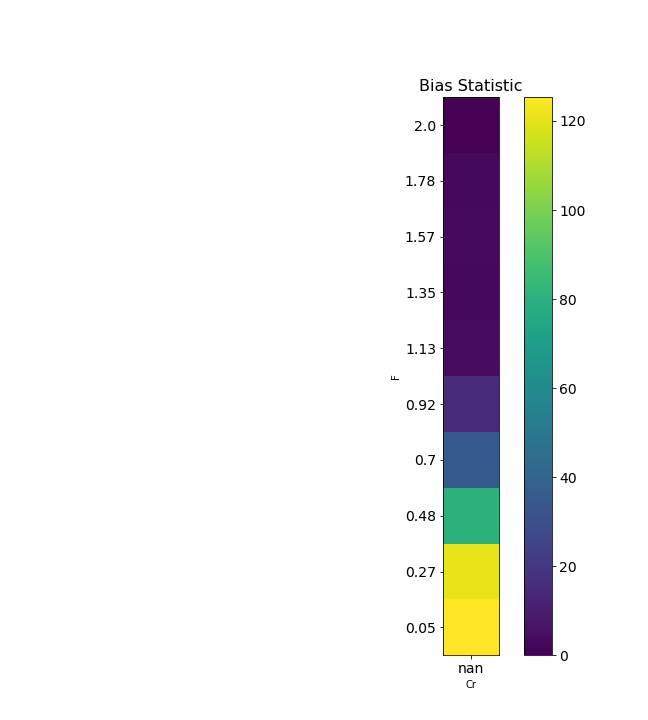}}
\subfloat[][\texttt{p100-sat}]{\includegraphics[width=.10\textwidth,trim=140mm 22mm 20mm 33mm,clip]{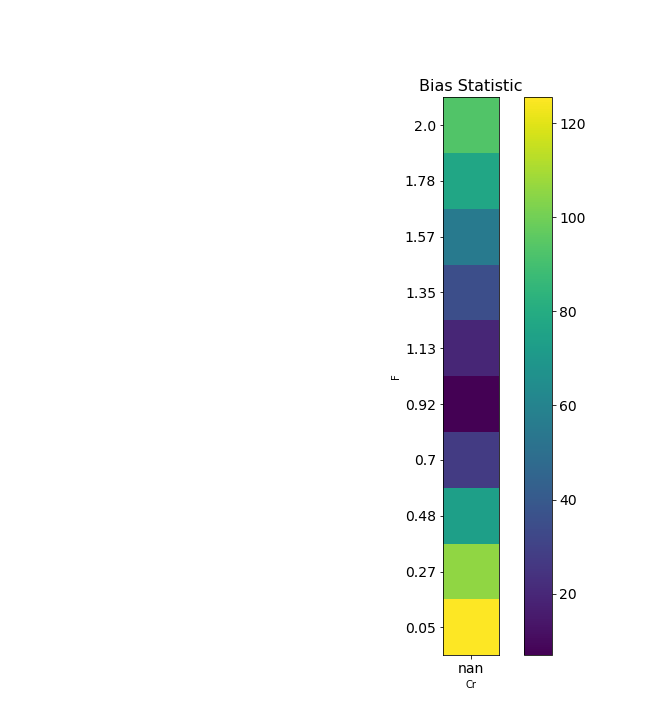}}
\subfloat[][\texttt{p100-tor}]{\includegraphics[width=.10\textwidth,trim=140mm 22mm 20mm 33mm,clip]{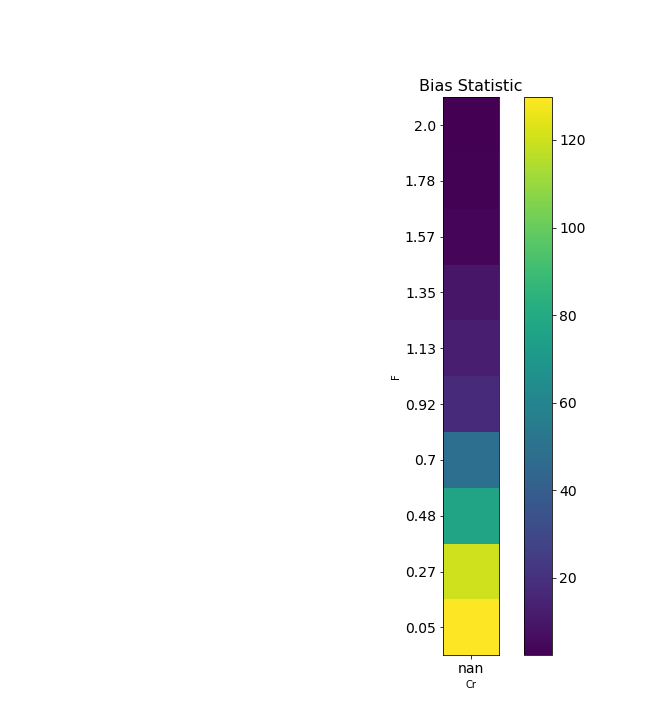}}
\caption{Bias indicator values for \texttt{DE/curr-to-rand/1} configurations with \texttt{p20} and \texttt{p100} and different values of $F$ (shown on the vertical axis) and strategies --- this mutation does not require a crossover and, thus, does not have $C_r$ parameter, see Section~\ref{sect:algorithms}. Note a different colour scale indicating the strength of SB.\label{fig:Crnan}}
\end{figure*}

\subsection{Analysis}
\paragraph{Effect of $F$ and $C_r$} 
Setting of DE parameters can greatly affect the bias, depending on the algorithm configuration. Over all the results, we can split the bias trends in three groups regarding $F$ and $C_r$ settings: 
\begin{itemize}
    \item Group 1 where the bias is dependent on both $F$ and $C_r$ and where increasing either of the two increases the bias statistic. 
    \item Group 2 where $F$ plays an important role in the bias, but $C_r$ does not. 
    \item Group 3 where only specific combinations of $F$ and $C_r$ seem to cause (mild) bias.
\end{itemize}

For example, in the first group, in Figure \ref{fig:DEbobsP5}, there is a clear upwards trend in the bias indicator when either $F$ or $C_r$ value increases. In Figure \ref{fig:DEboesP100} we can observe an increase in bias when $F$ increases, but the bias is only occurring for high $C_r$ settings (group 2). For some algorithm configurations, specific (lower) $F$ settings cause bias, such as in Figure \ref{fig:DEbobtP5} and Figure \ref{fig:DEbobuP20} (group 3). The bias in group 3 occurs mostly around $C_r = 0.76$ and $F = 0.48$.  There are also configurations that do not show any clear bias pattern for $F$ and $C_r$ settings, such as in Figure \ref{fig:DEbobd_P100}

In the cases of \texttt{DE/curr-to-rand/1} there is no crossover operator, but we can see clear trends in different $F$ values and different correction strategies (Figure \ref{fig:Crnan}). For example, when using a \texttt{COTN} strategy low $F$ settings cause considerable bias, while very high $F$ settings only cause mild bias. For \texttt{DE/curr-to-rand/1} with saturation strategy we can observe that both low and high $F$ values cause strong bias and only $F=0.92$ shows little bias (Figure \ref{fig:p20sat}).

For the following algorithms (Group 1), only low $C_r$ and $F$ settings ($< 0.48$) should be used to avoid bias
(all \texttt{sat}-configurations, irrespective of population size):
\begin{itemize}
    \item \texttt{DE/rand-to-best/bin}, 
    \item \texttt{DE/rand-to-best/exp}, 
    \item \texttt{DE/rand/2/bin}, 
    \item \texttt{DE/rand/2/exp},
    \item \texttt{DE/curr-to-best/1/bin},
    \item \texttt{DE/curr-to-best/1/exp}, 
    \item \texttt{DE/best/1/bin}, 
    \item \texttt{DE/best/2/bin}, 
    \texttt{DE/best/2/exp},
    \item \texttt{DE/best/1/exp}.
\end{itemize}

\paragraph{Effect of population size}
A larger population size slightly increases the structural bias as can be seen for example in Figure \ref{fig:Crnan} when comparing \texttt{p20-mir} and \texttt{p100-mir}. This behaviour is consistent over all configurations that show mild or strong bias. See also Figures~\ref{fig:DEbobsP5}, \ref{fig:DEbobsP20}, \ref{fig:DEbobsP100}.

\paragraph{Effect of mutation} In comparison with the other algorithm configuration settings, the choice of mutation operator has only little effect on the structural bias. Over all configurations, we can say that \texttt{rand-to-best/2} shows a slightly stronger bias than the other mutation operators.

\paragraph{Effect of crossover} Crossover has a strong effect on the bias, we see that for \texttt{bin} crossover the bias statistic is consistently higher than for \texttt{exp} crossover. The $F$ and $C_r$ settings also have a different influence when using either binary or exponential crossover as can be seen in Figures~\ref{fig:DEbobsP100}, \ref{fig:DEboesP100}

\paragraph{Effect of strategy} The correction strategy has the \textit{strongest} influence on the bias statistic. We can observe that \texttt{sat} strategy causes overall the most bias by far (Figures~\ref{fig:DEbobsP5}, \ref{fig:DEbobtP5} and Figures~\ref{fig:DEbobsP20}, \ref{fig:DEbobuP20}). Also strategy mirror and uniform cause strong bias in combination with mutation operator \texttt{current-to-rand/1}.  
 
\subsubsection{On recommended parameters settings}
Accumulated experience of heuristic optimisation community has resulted in a number of publications that provide typical DE parameter settings for the use by general practitioners: 
\begin{itemize}
    \item $F\in[0.5,0.9]$ and $C_r\in [0.8,1]$, $p=10n$, where $n$ is problem dimensionality \cite{tuningLampinen};
    \item $F\in\{0.5,0.9\}$, $C_r\in\{0.1,0.5,0.9\}$ and $p=50$ \cite{DasS11};
    \item \cite{bib:Zaharie2002} has instead investigated the inter-relationship among parameters $F$ and $Cr$, and discovered an optimal tuning of the crossover rate as a function of the scale factor;
\end{itemize}
 
We have selected values from our algorithmic setup closest to these recommendations ($F\in\{0.48,0.92\}$, $C_r\in\{0.05,0.52,0.999\}$, $p\in\{20,100\}$) and ranked all considered configurations according the strength of structural bias -- see Top 5s of such ranking in  
Tables~\ref{table:bias_cotn_dis_mir}, \ref{table:bias_sat_tor_uni}, per strategy. Since \texttt{DE/curr-to-rand/1} does not require mutation, we consider it separately, see Table~\ref{table:bias_ctr}.

The general conclusion of this exercise is that considered configurations with \texttt{COTN}, \texttt{dis}, \texttt{mir} and recommended parameter setting, result in at most mild structural bias which is attained for not-necessarily uttermost values of recommended parameters. Meanwhile, as stated previously, \text{sat}, \texttt{tor} and \texttt{uni} generally result in a stronger SB. For recommended parameter settings, these strategies deliver at most strong SB for cases of \texttt{DE/rand/2/$\star$}, \texttt{DE/best/$\star$} and \texttt{DE/curr-to-best/1/$\star$}, for \text{sat}, \text{tor} and \text{uni}, respectively. The case of \texttt{DE/curr-to-rand/1}, also appears to be strongly biased with \texttt{COTN} and \texttt{uni}.

Thus, general practitioners should avoid specific recommended parameter values if they choose \text{sat}, \texttt{tor} and \texttt{uni} strategies. The remaining strategies, should be used with care due to a possibility of mild SB.

\begin{table}
\caption{Top 5s of highest structural bias statistic scores among configurations with \texttt{COTN}, \texttt{dis} and \texttt{mir} strategies and parameter values in $F\in\{0.48,0.92\}$, $C_r\in\{0.05,0.52,0.999\}$, $p\in\{20,100\}$ closest to the recommended values.}\label{table:bias_cotn_dis_mir}
\centering
\begin{tabular}{|l|lll|r|}\toprule
\textbf{configuration} & \textbf{p} & \textbf{F} & $\mathbf{C_r}$ & \textbf{score} \\ \midrule
\texttt{DE/curr-to-best/1/exp-COTN} & 100 & 0.483 &	0.99 &	26.36\\
\texttt{DE/curr-to-best/1/bin-COTN} & 20 & 0.483 &	0.99 &	25.86\\
\texttt{DE/curr-to-best/1/bin-COTN} & 20 & 0.483 &	0.52 &	24.98\\
\texttt{DE/curr-to-best/1/bin-COTN} & 100 & 0.483 &	0.52 &	22.66\\
\texttt{DE/curr-to-best/1/exp-COTN} & 20 & 0.483 &	0.99 &	20.68\\
\midrule
\texttt{DE/curr-to-best/1/bin-dis} & 100 &	0,916 &	0,99 &	4,67 \\
\texttt{DE/rand-to-best/2/exp-dis} & 20 &	0,916 &	0,05 &	4,46 \\
\texttt{DE/rand/2/bin-dis} & 20 &	0,916 &	0,99 &	4,37 \\
\texttt{DE/best/1/exp-dis} & 20 &	0,483 &	0,05 &	4,15 \\
\texttt{DE/rand/2/bin-dis} & 20 &	0,483 &	0,99 &	4,08 \\
\midrule
\texttt{DE/best/2/exp-mir} & 20 &	0.483 &	0.99 &	5.44 \\
\texttt{DE/rand/2/bin-mir} & 100 &	0.483 &	0.52 &	4.84 \\
\texttt{DE/best/1/bin-mir} & 20 &	0.916 &	0.05 &	4.66 \\
\texttt{DE/best/2/exp-mir} & 100 &	0.916 &	0.52 &	4.61 \\
\texttt{DE/rand-to-best/2/exp-mir} & 100 &	0.483 &	0.52 &	4.11 \\
\bottomrule
\end{tabular}
\end{table}

\begin{table}
\caption{Top 5s of highest structural bias statistic scores among configurations with \texttt{sat},  \texttt{tor} and \texttt{uni} strategies and parameter values in $F\in\{0.48,0.92\}$, $C_r\in\{0.05,0.52,0.999\}$, $p\in\{20,100\}$ closest to the recommended values.}\label{table:bias_sat_tor_uni}
\centering
\begin{tabular}{|l|lll|r|}\toprule
\textbf{configuration} & \textbf{p} & \textbf{F} & $\mathbf{C_r}$ & \textbf{score} \\ \midrule
\texttt{DE/rand/2/exp-sat} & 20 & 0.916 & 0.99 & 174.85 \\
\texttt{DE/rand/2/bin-sat} & 100 & 0.916 & 0.99 & 171.71 \\
\texttt{DE/rand/2/exp-sat} & 100 & 0.916 & 0.99 & 171.60 \\
\texttt{DE/rand/2/bin-sat} & 20 & 0.916 & 0.99 & 165.99 \\
\texttt{DE/rand/2/bin-sat} & 20 & 0.916 & 0.52 & 163.24 \\
\midrule
\texttt{DE/best/1/bin-tor} & 20 & 0.483 & 0.99 & 21.23 \\
\texttt{DE/best/2/bin-tor} & 20 & 0.483 & 0.99 & 20.11 \\
\texttt{DE/best/1/exp-tor} & 20 & 0.483 & 0.99 & 15.15 \\
\texttt{DE/best/2/bin-tor} & 100 & 0.483 & 0.99 & 13.51 \\
\texttt{DE/best/2/exp-tor} & 20 & 0.483 & 0.99 & 12.04 \\
\midrule
\texttt{DE/curr-to-best/1/exp-uni} & 20    &	0.483 & 0.99 &	35.27 \\
\texttt{DE/curr-to-best/1/bin-uni} & 20    &	0.483 &	0.99 &	32.20 \\
\texttt{DE/curr-to-best/1/bin-uni} & 20    &	0.483 &	0.52 &	31.93 \\
\texttt{DE/curr-to-best/1/bin-uni} & 100   &	0.483 &	0.99 &	31.27 \\
\texttt{DE/curr-to-best/1/exp-uni} & 100   &	0.483 &	0.99 &	24.15 \\
\bottomrule
\end{tabular}
\end{table}

\begin{table}
\caption{Top 5s of highest structural bias statistic scores among configurations with \texttt{curr-to-rand} mutation and parameter values in $F\in\{0.48,0.92\}$, $p\in\{20,100\}$ closest to the recommended values.}\label{table:bias_ctr}
\centering
\begin{tabular}{|l|ll|r|}\toprule
\textbf{configuration} & \textbf{p} & \textbf{F} & \textbf{score} \\ \midrule
\texttt{DE/curr-to-rand/1-uni} & 100 & 0.483 & 91.60 \\
\texttt{DE/curr-to-rand/1-COTN} & 100 & 0.483 & 87.86 \\ 
\texttt{DE/curr-to-rand/1-COTN} & 20 & 0.483 & 87.08 \\ 
\texttt{DE/curr-to-rand/1-uni} & 29 & 0.483 & 81.74 \\
\texttt{DE/curr-to-rand/1-mir} & 100 &	0.483 & 79.23 \\
\bottomrule
\end{tabular}
\end{table}

\subsubsection{Comparison with previous results}
It is worth noting that configurations labelled biased with the graphical `parallel coordinate' approach in \cite{Caraffini2019} for a fixed parameter setting, have now been confirmed to be biased. An interesting case is \\ \texttt{DE/current-to-best/1/bin} with \texttt{dis} for all three population size. Indeed, it appears to be mildly biased via visual inspection \cite{Caraffini2019} while our tests confirm the formation of a high structural bias. 

\section{Emergence of structural bias in time}\label{sect:emergence_time}
Next to the emergence of SB in the parameter space, we have analysed a subset of algorithm configurations to investigate how the bias emerges over time, e.g. how the bias grows or shrinks after each evaluation.

For such analysis, each algorithm configuration is run for $300000$ fitness evaluations and $100$ runs (with different random seeds). We track SB by calculating the bias statistic using the active population of all $100$ runs for every evaluation.

\begin{figure*}
\centering
\subfloat[][\texttt{DE/best/1/bin-p5-sat} \newline $F=0.916,  C_r=0.99$]{\includegraphics[width=.33\textwidth,trim=6mm 8mm 20mm 17mm,clip]{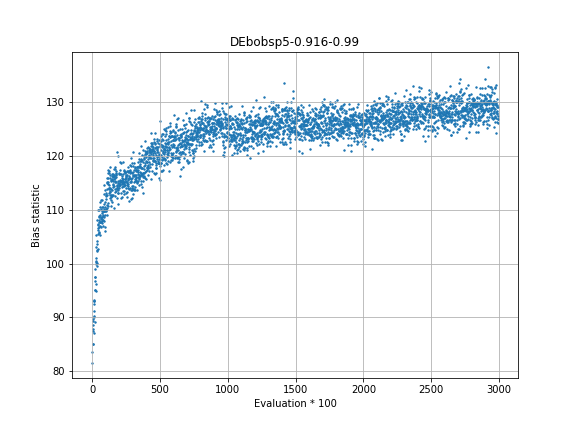}\label{fig:timeDEbobsP5}}
\subfloat[][\texttt{DE/best/1/bin-p20-sat} \newline $F=0.916, C_r=0.99$]{\includegraphics[width=.33\textwidth,trim=6mm 8mm 20mm 17mm,clip]{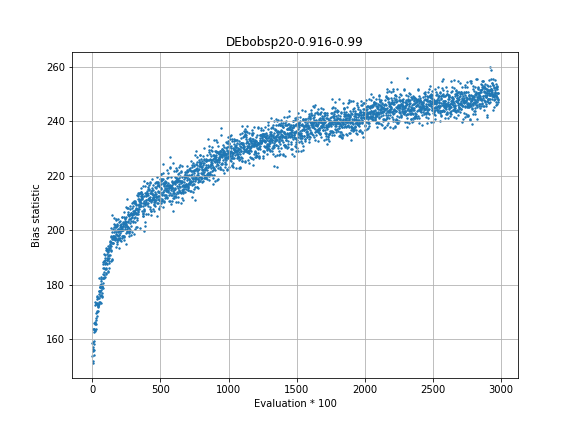}\label{fig:timeDEbobsP20}}
\subfloat[][\texttt{DE/best/1/bin-p100-sat} \newline $F=0.916, C_r=0.99$]{\includegraphics[width=.33\textwidth,trim=6mm 8mm 20mm 17mm,clip]{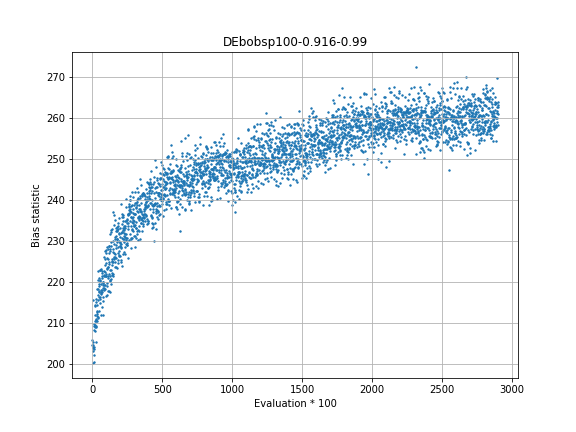}\label{fig:timeDEbobsP100}}\\ 
\vspace{-3mm}
\subfloat[][\texttt{DE/best/1/exp-p5-sat} \newline $F=2.0, C_r=0.99$]{\includegraphics[width=.33\textwidth,trim=6mm 8mm 20mm 17mm,clip]{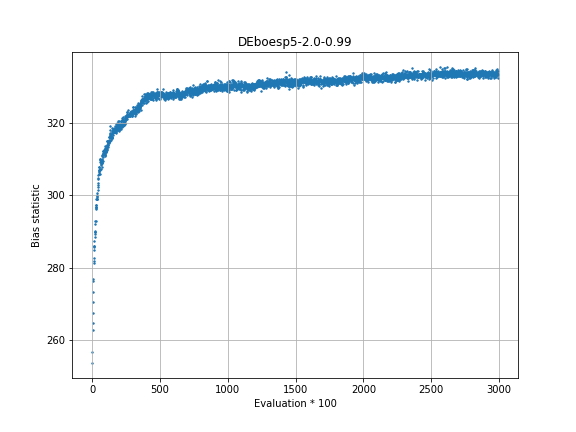}\label{fig:timeDEbobtP5}}
\subfloat[][\texttt{DE/current-to-best/1/bin-p5-sat} \newline $F=0.96, C_r=0.52$]{\includegraphics[width=.33\textwidth,trim=6mm 8mm 20mm 17mm,clip]{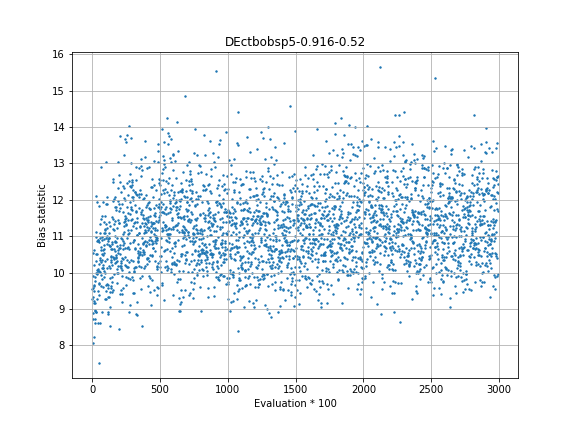}\label{fig:timeCTbobsp5}}
\subfloat[][\texttt{DE/rand/2/bin-p5-sat} \newline $F=1.78,  C_r=0.99$]{\includegraphics[width=.33\textwidth,trim=6mm 8mm 20mm 17mm,clip]{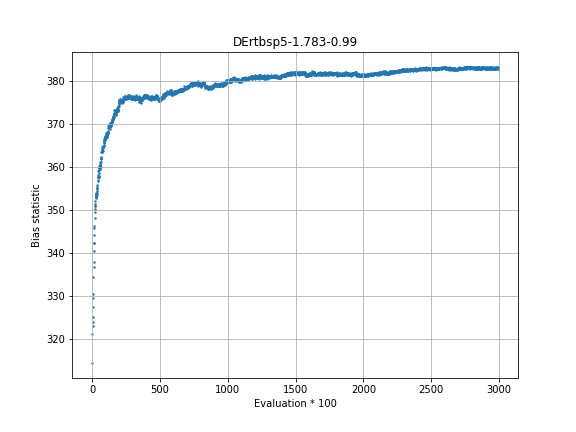}\label{fig:timeDEboesP100}}
\caption{Examples of bias emerging over time (fitness evaluations) for different algorithm configurations.\label{fig:time}}
\end{figure*}

\subsection{Analysis}
\begin{figure*}
\centering
\subfloat[][\texttt{DE/rand/2/bin sat p5} variations of $F$ and $C_r$]{\includegraphics[width=.49\textwidth,trim=10mm 5mm 20mm 17mm,clip]{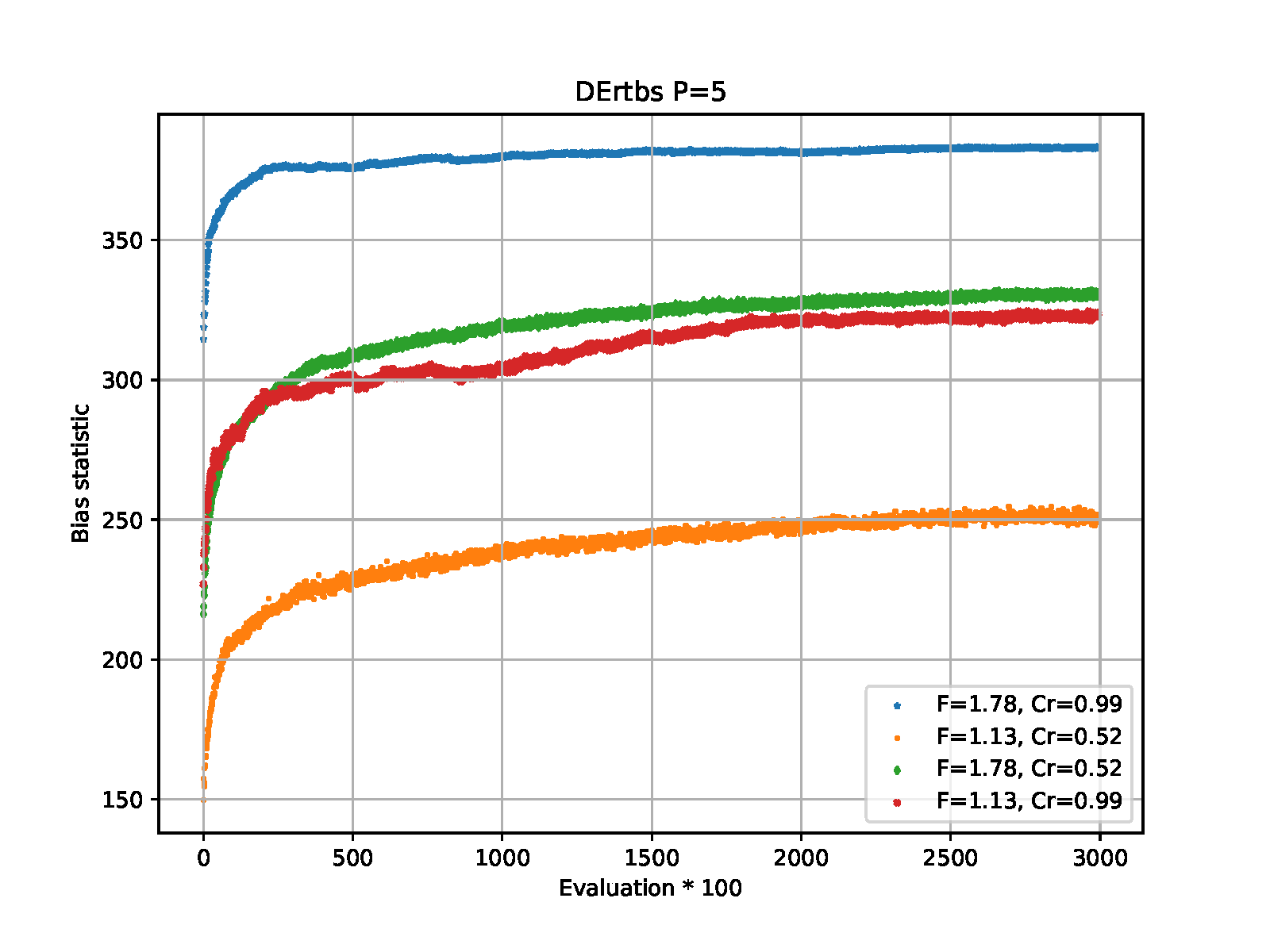}\label{fig:combiFCr}}
\subfloat[][\texttt{DE/rand/2/bin sat} variations of population size, $F=1.13, C_r=0.99$]{\includegraphics[width=.49\textwidth,trim=10mm 5mm 20mm 17mm,clip]{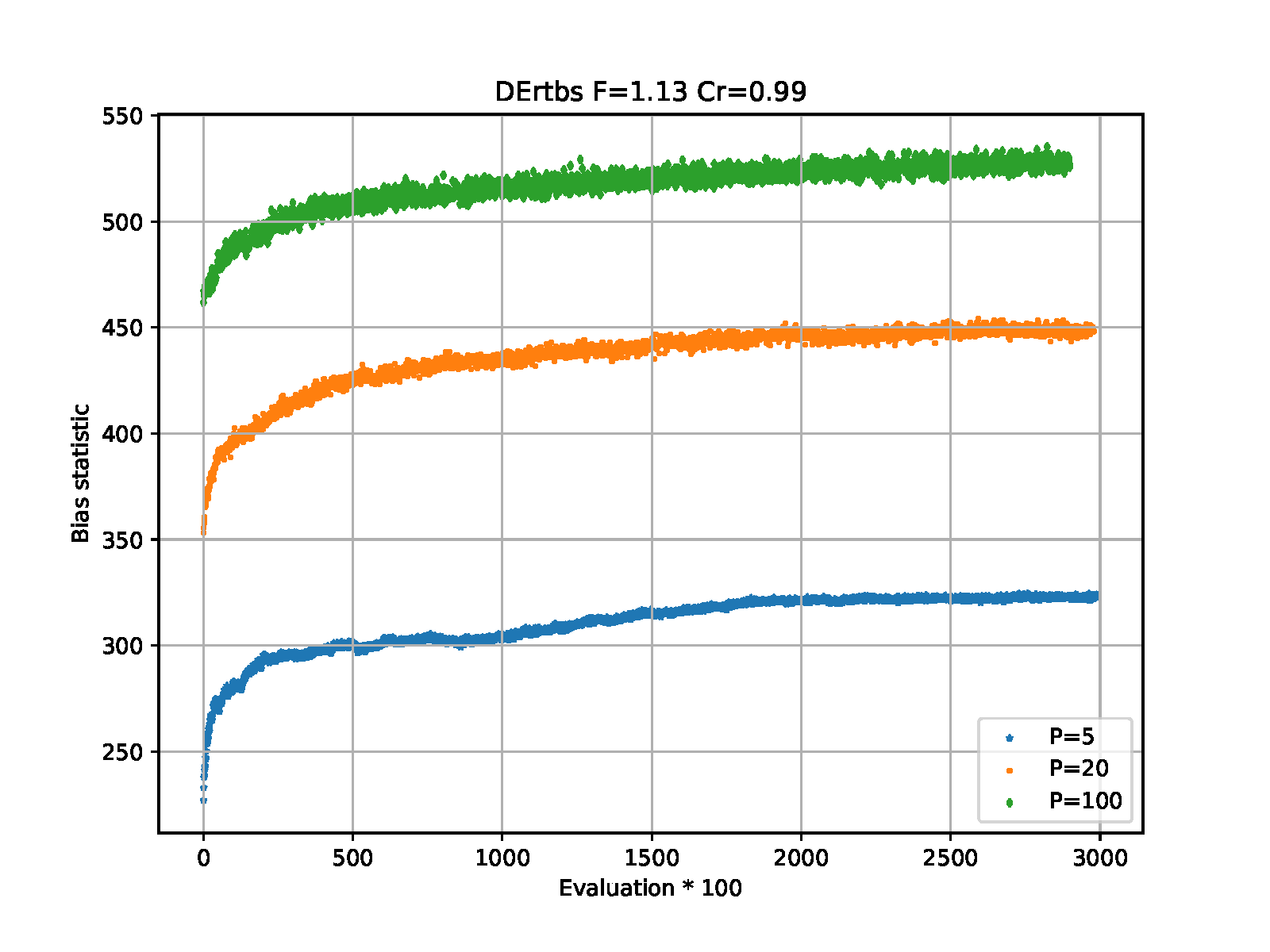}\label{fig:combiP}}
\caption{Examples of bias emerging over time (fitness evaluations) for different population sizes of DErtbs in Figure \ref{fig:combiP} and different $F$ and $C_r$ settings (Figure \ref{fig:combiFCr}).}
\end{figure*}
In Figure \ref{fig:time} we can see the bias emerging over time for a selection of $6$ different algorithm configurations and population sizes. The selected algorithms use \texttt{sat} strategy, as these algorithms clearly show the most bias and are therefore of the highest interest to analyse. 

We can observe that in all cases the algorithm starts unbiased. This is expected, since the initial population should be randomly uniform. After only a few evaluations though, the bias becomes evident and quickly climbs following a logarithmic curve. Even for cases with less bias, such as the \texttt{DE/current-to-best/1/bin-p5-sat} (Figure \ref{fig:timeCTbobsp5}), there is a clear trend starting at 0 and growing to an average of $11$ SB (i.e. the lower end of strong SB, following the definition above). In all plots we can also clearly see some degree of noise (roughly the same degree of noise in all cases, note that y-axes scale is different per figure). This noise is likely due to the stochastic nature of the experiment. It is also clear that if statistical bias manifests itself at some point in time, it does not disappear in the subsequent stages of evolution

When analysing different $F$ and $C_r$ settings over time for a configuration with saturation strategy, we can observe similar trends over time (Figure \ref{fig:combiFCr}). Higher $F$ and $C_r$ settings increase the bias reached over time. The slope of the curves however is only slightly affected. It is also clear that in this case both $F$ and $C_r$ have a similar importance for the bias statistic, increasing one and decreasing the other proportionally keeps the final bias stable. When we look at different population sizes (Figure \ref{fig:combiP}), we can observe a clear difference between sizes $5$, $20$ and $100$. This can be explained by the fact that the statistical test is performed on larger sample sizes for bigger populations ($p \times 100$), and does not necessarily mean that large populations also contribute to more bias.

\section{Conclusions and future work}\label{sect:conclusion}
The total of $10980$ DE algorithm configurations with various parameter settings have been analysed for presence of structural bias. It has been shown that a significant number of these configurations (almost $1$ in $5$) show strong structural bias, indicated by an Anderson-Darling based statistical test (bias statistic $> 10$). With many DE configurations show strong bias even within the ranges of originally proposed parameter settings $F$ and $C_r$. The saturation strategy stands out as one of the parameters that cause most bias, but also configurations like \texttt{DE/best/1/bin-tor} and \texttt{DE/curr-to-best/1/exp-uni} show significant structural bias. \\
Both $F$ and $C_r$ settings influence the emergence of structural bias. Generally speaking, high $F$ and $C_r$ values cause more bias than low values, however there are exceptions like the \texttt{DE/curr-to-rand/1}, where $F=0.05$ causes most structural bias in all cases.

Next to the effects of different configuration settings, the emergence of bias is measured over time (fitness evaluations). From the observations on these experiments it can be concluded that each algorithm configuration initially starts unbiased (with an initial population), but once structural bias starts to appear in the population it will only grow stronger during the remaining evaluations. 

Our future research will concentrate on a finer analysis of structural bias in DE with an improved measure to find other less evident types of bias, with a smaller sample size. Moreover, we will continue looking for factors that give rise to SB. 

\begin{acks}
We would like to thank Dr Hao Wang (LIACS) for providing the source code for bias measure computation and various fruitful discussions on the matter.
\end{acks}

\appendix
\end{document}